# Review on Optical Image Hiding and Watermarking Techniques

Shuming Jiao[a,c], Changyuan Zhou[a], Yishi Shi[b], Wenbin Zou[a], Xia Li[a]

[a]Shenzhen Key Lab of Advanced Telecommunication and Information Processing, College of Information Engineering, Shenzhen University, Shenzhen, Guangdong, China

[b]College of Materials Science and Opto-Electronic Technology, University of Chinese Academy of Sciences, Beijing 100049, China

[c]Tsinghua Berkeley Shenzhen Institute (TBSI), Shenzhen, 518000, China

E-mail: albertjiaoee@126.com

**Abstract**

Information security is a critical issue in modern society and image watermarking can effectively prevent unauthorized information access. Optical image watermarking techniques generally have advantages of parallel high-speed processing and multi-dimensional capabilities compared with digital approaches. This paper provides a comprehensive review on the research works related to optical image hiding and watermarking techniques conducted in the past decade. The past research works are focused on two major aspects, various optical systems for image hiding and the methods for embedding optical system output into a host image. A summary of the state-of-the-art works is made from these two perspectives.

## 1. Introduction

With the rapid development of Internet and information technology, the problem of unauthorized acquisition, transmission, manipulation and distribution of digital content is increasingly more severe in recent years. The research on information security has attracted considerable attention. In addition to digital approaches [1], optical information security approaches including optical image encryption, authentication and watermarking are extensively investigated [2,3] in the past decade. Optical approaches generally have advantages of parallel high-speed processing and multidimensional capabilities.

Image watermarking (or hiding) is a technique to embed a hidden image into another carrier image in a way that the hidden image is not accessible and perceptible by unauthorized users. The hidden image is generally referred to as watermark and the carrier image is generally referred to as host (or cover) image. In image watermarking (either implemented digitally or optically), there are usually several necessary criteria: (1) The host image shall not be significantly degraded after the watermark is embedded; (2) The watermark is imperceptible from the watermarked host image; (3) The watermark is robust and not easily removed or damaged from the host image. The watermark is tolerant to different types of attacks such as JPEG compression, cropping, rotation, scaling, noise, filtering and blurring. It shall be noted that this criterion is only applicable to robust watermark, rather than fragile watermark; (4) The watermark is sufficiently secure and difficult for an unauthorized user to illegally access.

Some comprehensive reviews on optical image encryption techniques [2,3] and on digital image watermarking techniques [4-8] are reported in the past. However, little work has been conducted on a survey of optical image watermarking techniques. The difference between encryption and watermarking (either optically or digitally) can be briefly described below. Encryption techniques transform the original image into noise-like ciphertext, which is not accessible by unauthorized users without the correct key. However, the existence of ciphertext is usually known to the third party. The aim of watermarking is to hide the existence of original image to unauthorized users.

Watermarking can be divided into two categories, i.e. robust watermarking and fragile watermarking. In the vast majority of works on optical image hiding, the watermark is defined as the former one. For robust watermarking, the watermark shall be intact when the host image is attacked and distorted. Since a robust watermark is very difficult to remove from a host image, it can be employed for applications such as copyright protection. On the other hand, for fragile watermark, once the host image is slightly modified, the watermark will be altered. A fragile watermark can be employed for verification applications such as integrity check of the host image. An intact fragile watermark indicates that the host image is in its original form without editing, damage and alternation.

The proposed optical image hiding or watermarking schemes in the past generally follow the framework illustrated in Fig. 1. In digital watermarking, the original watermark image is usually directly embedded into the host image with certain algorithms. However, in optical image watermarking, the original watermark image is usually employed as an input image to an optical system such as Double Random Phase Encoding (DRPE) system, off-axis holography system, phase shifting holography system, cascaded phase only mask architecture, Joint Transform Correlator (JTC), ghost imaging system and ptychography system. Then the system output (e.g. a hologram), named as "ciphertext" in this review article, is embedded into the host (or cover) image either optically or digitally. The embedding operations range from simple weighted addition to complicated adaptive signal embedding. It shall be noted that the host image is not necessarily a photograph but also other types of optical images such as a digital hologram. Finally, the original watermark image can be retrieved from the watermarked host image optically (or opto-digitally).

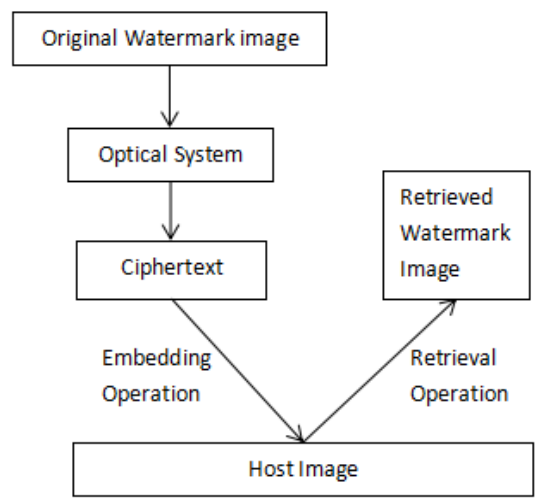

**Fig. 1.** General framework for optical image hiding (or watermarking) schemes

This paper presents a comprehensive review on various optical image watermarking systems and schemes proposed in the past decade. In Section 2, a review is conducted on common optical systems employed for image hiding. In Section 3, a review is conducted on the methods to embed optical system output into a host image. In Section 4, a brief conclusion is provided.

## 2. Various optical systems for image hiding and watermarking

A wide variety of optical systems have been attempted for optical image hiding or watermarking in the past works such as Double Random Phase Encoding (DRPE), off-axis holography system, phase shifting holography system, cascaded phase only mask architecture, Joint Transform Correlator (JTC), ghost imaging system and ptychography.

*2.1. Optical watermarking with Double Random Phase Encoding (DRPE)*

Double Random Phase Encoding (DRPE) scheme was first proposed by Refregier and Javidi in 1995 [9] and it has been extensively employed in optical information security area, including optical image hiding. DRPE can be expressed as Eq. (1).

$$g(x,y) = IFT\{FT[f(x,y)exp(j2\pi p(x,y))]exp(j2\pi q(u,v))\} \tag{1}$$

where $f(x,y)$ denotes the hidden image, $FT$ denotes Fourier transform, $IFT$ denotes inverse Fourier transform, $exp(j2\pi p(x,y))$ and $exp(j2\pi p(x,y))$ are two random-phase masks in spatial domain and frequency domain, and $g(x,y)$ denotes the ciphertext image (a complex signal) generated for the original hidden image.

In [10,11], a straightforward way to apply DRPE in image watermarking is proposed. The original watermark image $f(x,y)$ is first encrypted by DRPE and the cipher-text $g(x,y)$ (a complex signal) is weight added to the host image $h(x,y)$ in spatial domain, shown in Eq. (2), where m is a weighting coefficient. The watermark can be retrieved when conventional DRPE decryption steps are applied to the watermarked image $w(x,y)$, shown in Eq. (3). The host image will contribute tolerable level of noise in the retrieved watermark image $f'(x,y)$. An example of watermarking results is illustrated in Fig. 2.

$$w(x,y) = h(x,y) + m \cdot g(x,y) \tag{2}$$

$$f'(x,y) = IFT\{FT[w(x,y)]exp(-j2\pi q(x,y))\}exp(-j2\pi p(x,y)) \tag{3}$$

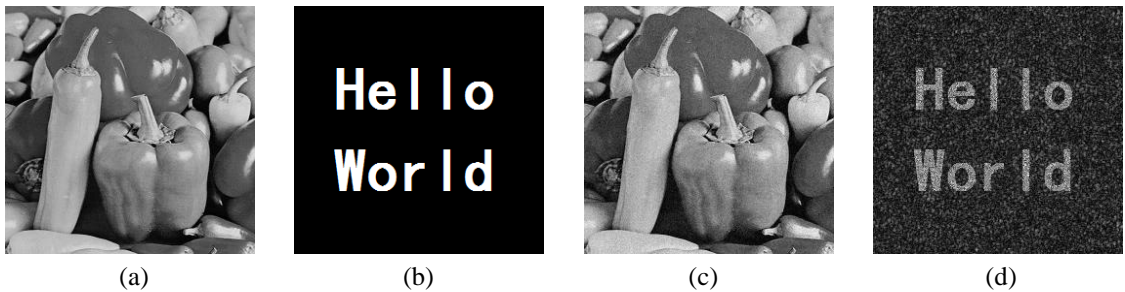

(a)          (b)          (c)          (d)

**Fig. 2.** An example of watermarking results from a DRPE scheme [10]: (a) host image $h(x,y)$; (b) watermark $f(x,y)$; (c) watermarked image $w(x,y)$; (d) retrieved watermark $f'(x,y)$.

The complex ciphertext after DRPE encryption can be quantized to a reduced number of discrete real values [12,13]. As a consequence, the data size of ciphertext is highly compressed and a reasonable recovered hidden image quality can still be maintained.

In addition to Fourier transform domain, the watermark can be encrypted by DRPE in Fractional Fourier transform (FrFT) domain. The fractional order parameter in FrFT offers extra security against attacks. The ciphertext is embedded into a conventional host image [14] or a Fresnel hologram [15]. Multiple watermarks can be encrypted by DRPE with different keys in FrFT domain and then multiplexed by superposition [16]. Each individual watermark can be retrieved separately from the watermarked host image. The original watermark image can be encrypted by other optical encryption schemes originating from DRPE such as phase-truncation asymmetric optical encryption [17].

The security strength of DRPE watermarking is analyzed and the possibility that such a watermark can be detected by a third party is discussed. A method to detect whether there is DRPE encrypted secret information embedded in a given image is proposed [18]. The lower-order bit planes of a host image containing DRPE encrypted watermark will exhibit stronger randomness than conventional natural images. From statistically analysis (t test), the existence of such watermark can be blindly detected.

A Double Random Phase Encoding (DRPE) system can hide a watermark image securely at fast processing speed. The original image can be retrieved by a DRPE system as well. However, a DRPE system is rather hard to implement in real optical experiments due to the precise alignment requirement of random phase masks. Most investigations on optical watermarking with DRPE are based on numerical simulation only. Another drawback of DRPE system is that the random phase masks will cause heavy speckle noise in the recovered watermark image.

*2.2. Optical watermarking with off-axis holography system*

Holography is a technique to record and reconstruct a three-dimensional (3D) scene on a two-dimensional (2D) plane (i.e. a hologram) based on optical diffraction and interference. A digital hologram stores viewing perspective and depth information of a 3D object. A 2D image (either complex or real pixel values) can be converted to a hologram as well. From information processing perspective, the local information in the object image is redistributed globally in the hologram image. Consequently, one portion of a hologram can contain information of the entire reconstructed image. Due to this property, one hologram has an inherent property to resist cropping, noise and distortion, which is suitable for robust watermarking applications. Hologram watermarking can be employed in practical applications such as ID card authentication [19] by embedding a hologram of security token data into the photograph on an ID card.

A hologram can be captured from the original watermark image with a holographic recording system such as off-axis holography system or phase shifting holography system. For example, in [20], the original watermark image is modulated by a random phase mask and then Fourier transformed. After that, an off-axis reference wave is superposed on the transformed result. The interference intensity pattern can be recorded an off-axis Fourier hologram of the watermark. The hologram is weight added to the spatial domain of a host image. The original watermark can be holographically reconstructed from the watermarked image directly. The optical setup for off-axis Fourier holography is presented in Fig. 3.

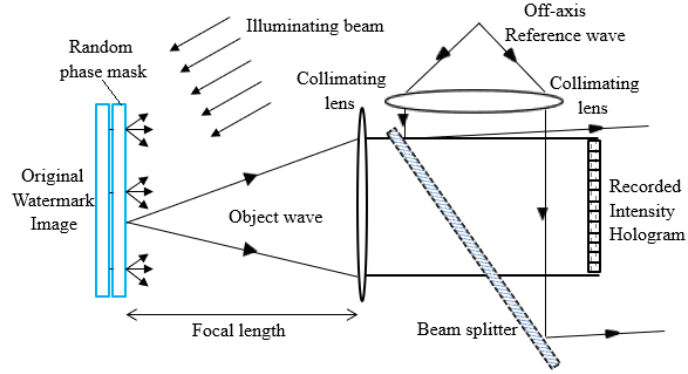

**Fig. 3.** An off-axis Fourier holography architecture.

*2.3. Optical watermarking with phase shifting holography system*

Phase shifting holography or phase shift interferometry (PSI) is an alternative approach for recording a complex light field generated from a 3D object, in addition to off-axis holography. In phase shifting holography, a complex light field of original watermark image is interfered with phase shifted reference wave and multiple intensity patterns (interferograms) are captured. The embedding of multiple interferograms into the host image allows the insertion of original watermark image information.

An example of the phase shifting interferometric architecture for image hiding [21-23] is illustrated in Fig. 4. In [22], the original watermark image is first encrypted by DRPE and then the encrypted light field is recorded as three interferograms (or phase shifting holograms). Each interferogram is weight added to a host image. By addition and subtraction operation of the three watermarked images, the host image can be fully cancelled out or at least partially suppressed and the encrypted light field can be reconstructed, followed by a recovery of original watermark image.

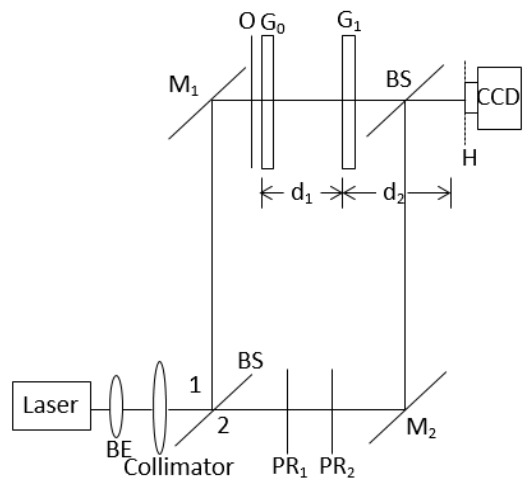

**Fig. 4.** A phase-shifting interferometry architecture for image watermarking. BE, beam expander; BS, beam splitter; PR: phase retarder; M, mirror; O, original watermark image; $G_0$, $G_1$, random phase masks.

In [24], the interferograms captured from multiple watermark images are multiplexed by superposition and each original watermark image can be retrieved with certain level of noise [24]. Two hidden images can be employed to modulate the amplitude and phase part of a complex object in PSI correspondingly to avoid crosstalk with each other [25].

The advantage of phase-shifting holography approach over off-axis holography is that the watermark can be perfectly retrieved with reduced noise or no noise caused by the host image. The extra expense is that multiple watermarked host images are required for one single hidden image. To overcome this limitation, dual channel simultaneous PSI can be employed and two interferograms instead of three are required for each watermark [26]. Compressive sensing can be employed to further reduce the data amount [26].

For both off-axis holography and phase shifting holography systems discussed in Section 2.2 and Section 2.3, the watermark image and host image are not necessarily two-dimensional but also three-dimensional. In [27], one hidden 3D object can be watermarked into another host 3D object indirectly with hologram to hologram watermarking. The digital hologram of a hidden object is encrypted by Double Random Phase Encoding (DRPE) and weight added to the digital hologram of host object, shown in Fig. 5. The hidden 3D object and host 3D object can be optically reconstructed by a DRPE system and holography system respectively.

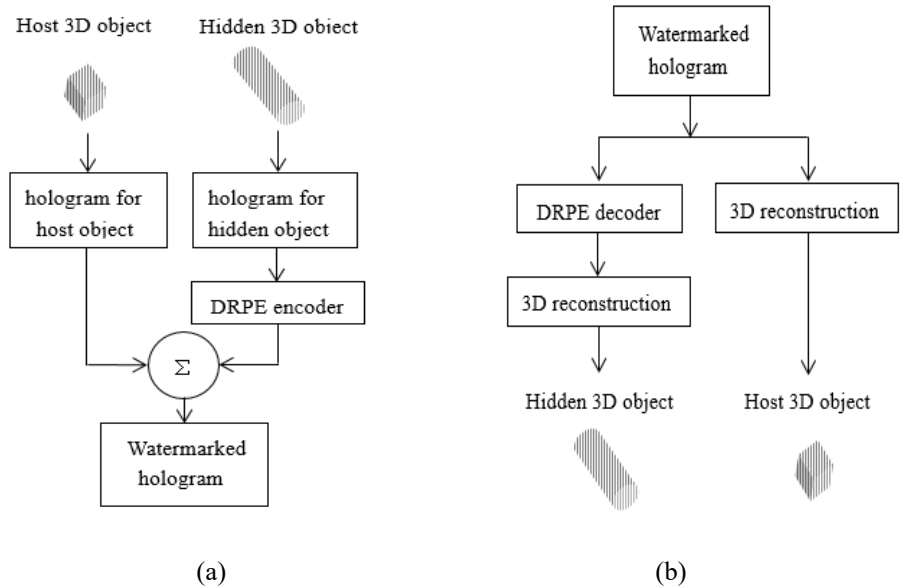

(a)          (b)

**Fig. 5.** 3D object watermarking scheme based on holography [27]: (a) watermark embedding; (b) watermark retrieval.

In [28], one micro image imperceptible to human eyes is embedded into a host image with much larger pixel size by hologram to hologram watermarking. Both the original watermark image and host image are converted to holograms of identical pixel size via scaled diffraction. Then one hologram is embedded into another by weighted addition.

*2.4. Watermarking with cascaded phase only mask architecture*

Watermarking can be implemented in a cascaded phase only mask architecture [29-35] with phase retrieval algorithms. This architecture differs from the general framework shown in Fig. 1 in a way that the hidden image is not directly embedded into the host image whereas the hidden image can be displayed when the host image and two correct phase only masks are placed in the cascaded phase only mask system.

An example of such an optical system is illustrated in Fig. 6. The two phase-only masks are generated by iterative projections onto constraint sets (POCS) algorithm, with the host image placed in the input plane and the hidden image planed at the imaging plane. When the host image and two corresponding computed phase masks are placed in the cascaded Fresnel transform system [29,30], the hidden image (watermark image) can be observed in the imaging result.

The computed phase mask and the host image can be further combined to form a concealogram [31]. In [32], the cascaded phase only mask system is employed for the watermarking of multiple images into one single host image. The different watermark images are displayed at different imaging distances in the system and two phase masks can still be generated with a modified iterative algorithm. In [33], a cascaded Fourier transform system with lens, rather than lensless Fresnel transform system, is attempted.

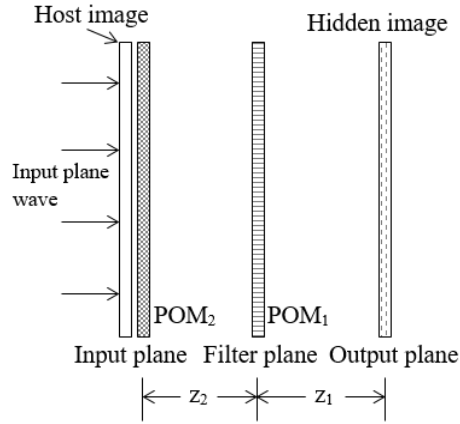

**Fig. 6.** A cascaded lensless Fresnel digital holography (CFDH) architecture [29,30] (POM, phase only mask).

In [35], the cascaded phase only mask architecture is implemented in optical gyrator transform domain. The original watermark image is first randomized by Arnold transform either optically or digitally and then encoded as the amplitude part in the output imaging plane of a gyrator transform system. The host image is set as the amplitude part of gyrator transform input plane. The phase masks are calculated by Gerchberg–Saxton iterative algorithm, similar to the other works stated above.

In [33], an image hiding technique is proposed that combines the complementary advantages of cascaded Fourier transform and RSA encryption algorithm, which is a hybrid of optical and digital security approaches. A random intensity image is placed in the input plane and the original watermark image is placed in the output imaging plane. Then two phase masks are obtained by cascaded iterative Fourier transform (CIFT) algorithm. The encrypted random intensity image by RSA algorithm and two retrieved phase masks are watermarked into a host image by lower-order bit replacement and weighted addition correspondingly.

As a unique feature, the cascaded phase only mask architecture does not require additional embedding operation to insert optical system output into the host image and the watermarking is non-destructive to the host image. On the other hand, the cascaded phase only mask architecture has a major drawback of difficult real optical implementation, similar to DRPE system. It is a very challenging task to align random phase-only masks precisely in an optical system.

*2.5. Watermarking with Joint Transform Correlator (JTC) architecture*

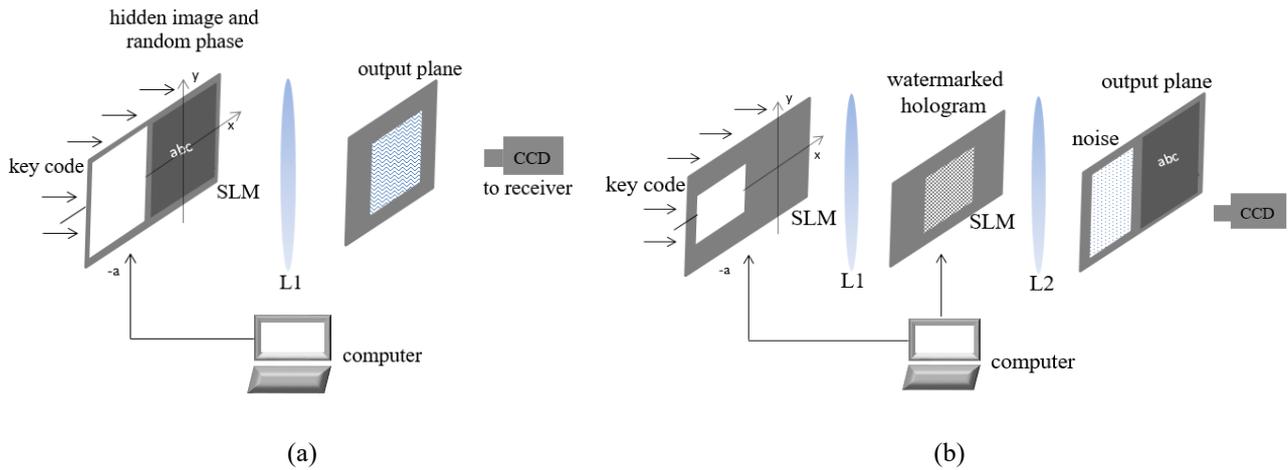

**Fig. 7** A Joint Transform Correlator (JTC) architecture for image watermarking [36] (SLM: spatial light modulator; $L_1$, $L_2$: lens): (a)watermarking embedding; (b)watermarking retrieval.

A joint transform correlator (JTC) architecture is employed for both watermark embedding and retrieval [36], illustrated in Fig. 7. The hidden image and a key code image are jointly placed in a JTC architecture and the former is encrypted. The encrypted information is weight added to the Fourier transformed hologram of host image. The hidden image can be retrieved from the watermarked hologram by JTC architecture with correct key code. In addition, a joint Fresnel transform correlator architecture can be employed for optical image watermarking [37].

The JTC architecture is relatively easy to implement experimentally and the optical processing speed is rather fast. However, unlike holography and ptychography, JTC only supports the watermarking of a 2D object image, instead of both 2D and 3D object image.

*2.6. Optical watermarking with ghost imaging system*

Ghost imaging is a single pixel imaging technique based on the correlation between object beam signals and reference beam signals. Two identical beams, the object beam and the reference beam, are generated from the same light source. The target object image is illuminated by the object beam and the light intensity is recorded by a single pixel bucket detector. In the meanwhile, the reference beam is directly recorded optically or digitally, without interacting with the target object. After a large number of beam pattern illuminations, the original object image can computationally reconstructed from the recorded intensities on the single pixel detector and corresponding reference patterns. A typical optical setup for ghost imaging is demonstrated in Fig. 8.

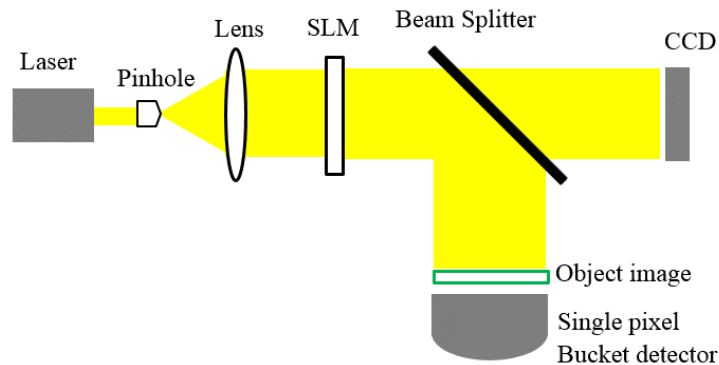

**Fig. 8.** Optical setup for a ghost imaging system.

A ghost imaging system can be employed for information security applications by employing the recorded intensity values as ciphertext and the reference patterns as key. The ghost imaging system output (intensity values recorded by the single pixel detector) can be embedded into a host image for watermarking application [38]. Furthermore, the ghost imaging output of one object image can be hidden into the ghost imaging output of another object image [39,40].

In all the other optical systems discussed in this paper, an image sensor array is required to record the ciphertext generated from the original watermark image. However, in a ghost imaging system, a simplest single pixel sensor is capable of recording the image information of 2D image or even 3D image. As a consequence, a ghost imaging system is suitable for watermarking an original object image in an environment where conventional image sensor array does not work. The cost of ghost imaging for watermarking is that the acquisition of original watermark image takes long time due to the repeated illumination with a large number of reference patterns.

*2.7. Watermarking with ptychography*

Ptychography is an optical technique to record the complex amplitude of one object by recording multiple diffraction intensity patterns of the object through partially overlapped probes (shown in Fig. 9). In contrast to holography, ptychography is a 3D imaging technique without reference wave and the complex information is recovered by phase retrieval algorithm from the multiple diffraction intensity patterns. The optical setup for a ptychography imaging system is illustrated in Fig. 9.

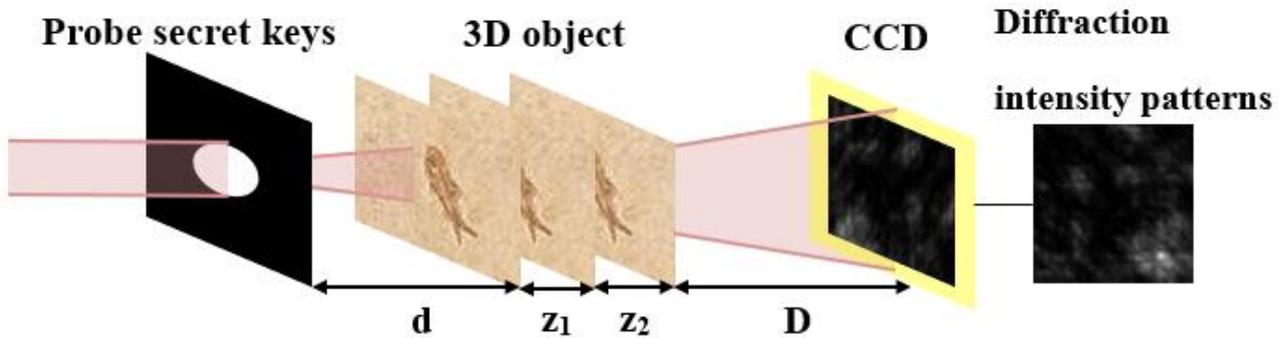

**Fig. 9.** Optical setup for a ptychography imaging system

Ptychography can be employed for the watermarking of a 3D object [41]. The multiple diffraction intensity patterns of a 3D object are recorded and weight added to host images. Each depth section of the 3D object can be reconstructed from the patterns extracted from host images. In this scheme, the crosstalk between different depth sections of the 3D object can yield a noise-like recorded diffraction patterns, which are favorable for watermarking applications. As a drawback, a large number of optical exposures, recorded diffraction patterns and host images are required. An improved single-shot-ptychography watermarking scheme is proposed by recording one diffraction intensity pattern constituting a series of tiny diffraction spots [42]. Multiple-image hiding with ptychography is also investigated recently [43].

As an advantage, ptychography is easier for optical experimental implementation without a strict precision requirement as in many other methods such as DRPE schemes.

*2.8. Comparison of various optical systems for image hiding*

The features of various optical image hiding systems stated above are summarized in Table 1, including their pros and cons.

**Table 1**
Comparison of various optical systems for image hiding

| Optical system for image hiding | Pros | Cons |
| --- | --- | --- |
| Double Random Phase Encoding (DRPE) | Double random phase masks as key to ensure security; Fast processing | Hard to implement in real optical experiment; Speckle noise due to random phase mask |
| Off-axis holography system | Support 3D object watermarking | Reference wave required |
| Phase shifting holography system | Support 3D object watermarking; Host image can be easily removed in watermark retrieval | Reference wave required; Multiple host images required |
| Cascaded phase only mask architecture | Non-destructive to host image | Hard to implement in real optical experiment |
| Joint Transform Correlator (JTC) | Easy to implement optically | Not applicable for watermarking 3D object |
| Ghost imaging system | Simplest image detector | Long image acquisition time; Noise in reconstructed image |
| Ptychography | Support 3D object watermarking; Noise resistant; Easy to implement optically | Recording of multiple diffraction patterns required |

These optical image hiding systems can be further improved from the following perspectives in future works.

(1) Some optical systems such as DRPE and cascaded phase only mask system are hard to implement in optical experiments due to the precise alignment requirement under coherent light illumination. This problem can be possibly tackled by combining these optical systems with some incoherent operations such as visual cryptography [44,45], where an approximate overlapping (rather than very precise alignment) of optical components can yield correct decrypted result.

(2) For ghost imaging system, the acceleration of image acquisition speed for watermarking can be possibly achieved by properly designing the reference illumination patterns [46] and image reconstruction algorithms [47]. The information security strength and imaging speed shall be ensured at the same time.

(3) For ptychography system, the data amount of system output is huge (multiple diffraction patterns). Compression method shall be developed to further reduce the data size while maintaining the retrieved watermark image quality.

(4) The parallel high-speed processing and multidimensional capabilities is a significant advantage of optical security systems. This feature can be better utilized for the watermarking of a huge number of images and video clips in future works.

## 3. Methods for embedding optical system output into host image

In the general framework for optical image hiding or watermarking, shown in Fig. 1, after the original watermark image is converted to be a ciphertext with optical systems described in Section 2, the ciphertext needs to be embedded into the host image in certain manner. In many previous works, the embedding operation is implemented with tailor-made digital algorithms based on the signal properties of optical system output and host image. The algorithms for embedding a hologram (generated from the original watermark image with a holography optical recording system) into a host photograph image are most extensively investigated in the past. Some of these algorithms are applicable for embedding other types of ciphertext signals to other types of host images as well. The following representative embedding methods have been reported in the past literatures.

### 3.1. Weighted addition in spatial domain and transformed domain

The most straightforward way of embedding an external signal into a host image is weighted addition in spatial domain like Eq. (2) and Fig. 10(a), employed in many works such as [10,11,20]. In [20], a hologram (as ciphertext) is weighted added into a host photograph. In [48], when the host image is a hologram (rather than a photograph) and the ciphertext is a DRPE system output, an optimal weighting factor criterion is derived by minimizing the summarized error of reconstructed object image of host hologram and decoded watermark. As a consequence, the image quality of host hologram and retrieved watermark can be both well preserved. In addition, the hologram of an original watermark can be superposed onto a host image with an appropriate weighting factor in a transformed domain (such as discrete cosine transform, discrete wavelet transform and singular value decomposition).

In [49], the hologram is weight added to the median-frequency coefficients of the discrete cosine transformed (DCT) host image instead of spatial domain, shown in Fig. 11(b). The low frequency and high frequency coefficients are well preserved so the host image quality after watermarking is better than the work [20]. The drawback in the work [20] is that the host image needs to be low-pass filtered first before the superposition of embedded hologram, can be overcome. In [34,50], hologram watermarking is performed in the discrete wavelet transform (DWT) domain. The hologram is embedded into the low-frequency wavelet coefficients of the host image by quantization index modulation (QIM). The wavelet coefficients are quantized by two different quantizers according to the binary bit values of the hidden holographic data. In [51], singular value decomposition (SVD) is applied to the host image and the hologram is embedded to the singular values of host image, which possesses good stability.

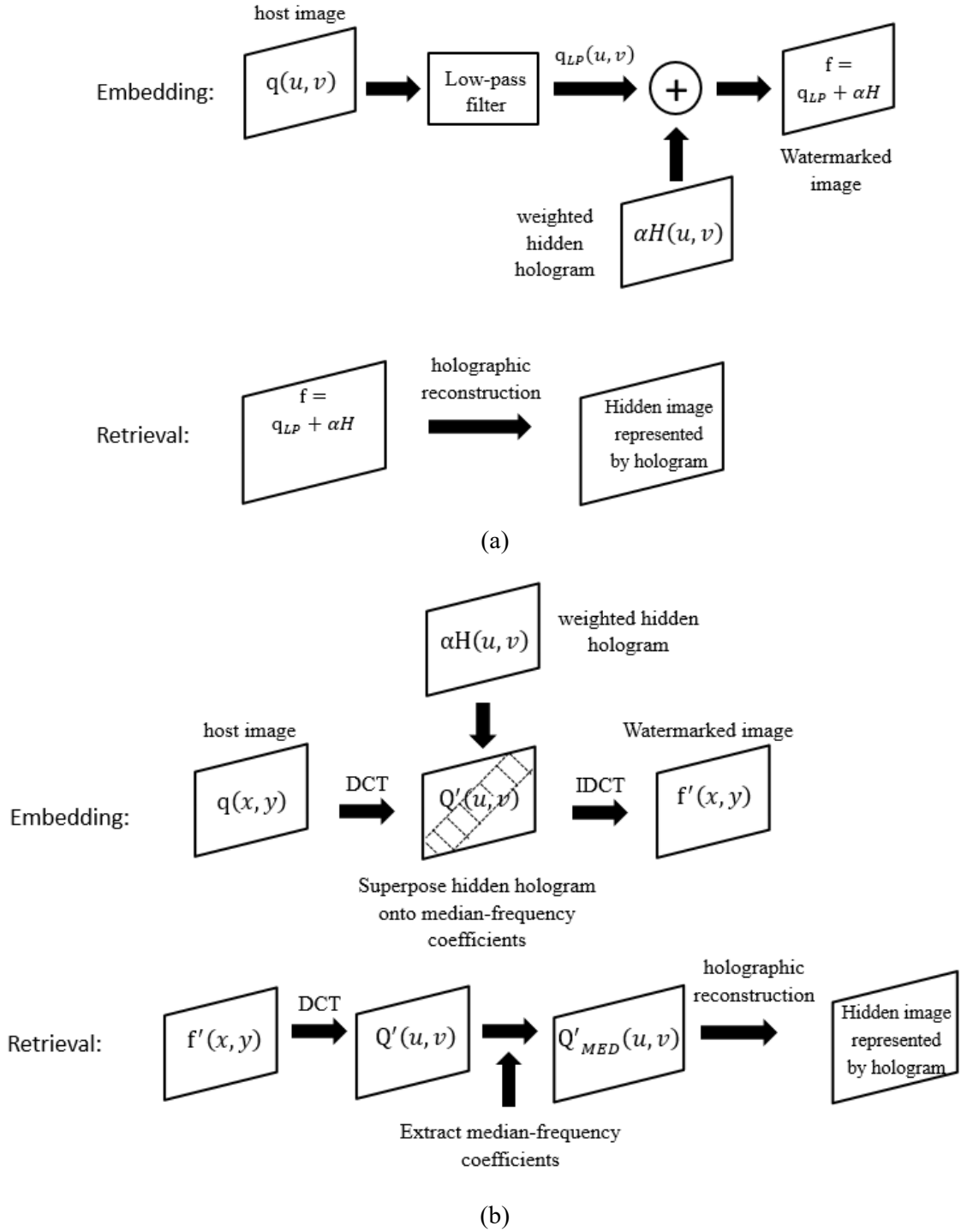

**Fig. 10.** Comparison of watermarking a hidden hologram into a host image

in (a) spatial domain; and (b) discrete cosine transformed domain.

The transform domain approaches [49-51] exhibits advantages over the spatial domain approach [20] in terms of image quality, robustness and stability. However, some additional digital operations are required in the watermark retrieval for these methods [49-51] and full optical watermark retrieval can be realized in the spatial domain method [20].

*3.2. Adaptive data embedding into host image*

Instead of using a fixed weighting coefficient [20,49], the embedding can be performed in an adaptive manner in spatial domain [52] or transformed domain [53,54] to better preserve host image quality. In [52], the number of lower-order bits replaced by embedded data in each pixel of the host image is proportional to the original pixel amplitudes in the host image. One hologram with optimized phase quantization levels for each pixel is calculated by iterative Gerchberg–Saxton method.

An adaptive watermarking scheme for embedding a gray-level hologram into the wavelet coefficients of a host image is proposed [53]. An improved Fuzzy C-Means Clustering with modified MAXMIN algorithm is utilized to find the most suitable blocks in the host image for embedding the hologram. In addition, a novel iterative algorithm is proposed for the embedding process to enhance the watermark robustness. This watermarking method demonstrates a distinct advantage in better robustness to a JPEG compression attack on the host image, compared with non-adaptove watermarking methods such as [49].

In [54], the hologram is watermarked into the low frequency sub-band of the contourlet-transformed host image by an improved quantization embedding algorithm. In addition, a particle swarm optimization algorithm is employed to optimize the parameters of the entire watermarking system to obtain a balance between watermarking imperceptibility and robustness.

These adaptive watermarking schemes can give more satisfactory watermarking performance (embedding more hidden ciphertext information into the host image while causing less damage to the host image) but the expense is much higher complexity.

*3.3 Enhanced resistance to certain attacks*

Some methods [55-58] are developed to further enhance the robustness and attack resistance capability of hologram watermarking methods such as resistance to geometric attacks and print-scan process.

The inverse log-polar mapping of a hologram, rather than the original hologram, can be watermarked into a host image [55] to better resist rotation and scaling attacks (shown in Fig. 11.). A method to resist strong cropping attack is proposed by embedding a Burch CGH watermark into Discrete Fourier Transform (DFT) domain of a host image [56]. A geometric distortion correction method [57] is proposed for a host image embedded with hologram data in the low-frequency subband wavelet coefficients. The geometric distorted watermarked host images are recovered by scale invariant feature transform (SIFT), the invariant centroid, and the pulse coupled neural network (PCNN). This scheme can resist common geometric attacks including rotation, scaling, translation, image flipping, occlusion and cropping.

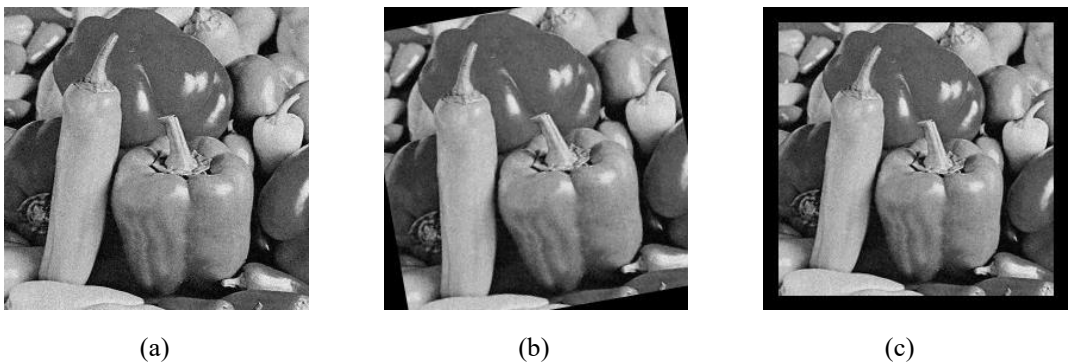

(a)          (b)          (c)

**Fig. 11.** Rotation and scaling attacks: (a) Original host image; (b) Rotated host image; (c) Scaled host image.

A hologram watermarking scheme is proposed to allow the host image survive in hardcopy print-scan process [58]. The CGH is generated after the original watermark image is conjugate-symmetric extended.

Then DCT spectrum of the CGH is inserted into the DCT spectrum of the host image.

In some proposed semi-fragile watermarking method, the embedded hologram is fragile to certain attacks while robust to other attacks [59]. For example, in [59], the original watermark image is first digitally encrypted with linear congruency method (LCM). Then Fourier-transformed hologram of the ciphertext is numerically generated and embedded in the cover image (such as a photograph). This scheme can monitor the integrity of a photograph and identify replaced, edited or damaged parts. At the same time, it is resistant to cropping, geometrical transformation and contrast modification to some extent.

*3.4. Watermarking in dithering and halftoning*

In addition to gray scale host images, a hologram can be embedded into a binary host image in the dithering or halftoning process.

Image dithering can convert a gray scale image into a binary image by allowing the density of binary pixels approximates the average gray levels. Watermarking can be implemented in the ordered dithering process of a host image [60]. A binary computer generated hologram (CGH) is generated from a watermark image by error diffusion method. Micro blocks of the host image are dithered with two different dithering matrices according to the binary pixel values of the CGH so that the binary CGH is watermarked.

Halftoning is an alternative technique for converting a gray scale image to a binary image by controlling the size and repetition frequency of an elementary binary dot pattern. A Lohmann type hologram with elliptical diffraction patterns is embedded into a halftoned host image by modulating the elementary halftoning pattern size [61,62]. In [63], the host image information is stored in the amplitude part of a complex wavefront and the watermark information is associated with the phase part. The amplitude and phase part of the complex wavefront is converted to a binary intensity image by dot-area modulation (DAM) halftoning and dot-position modulation (DPM) halftoning respectively. The watermark information contained in the binary image can be revealed with an optical correlator and appropriate key function.

*3.5. Digital watermarking for holograms*

This review paper is mainly focused on optical image hiding or watermarking techniques described by the framework shown in Fig. 1. In some past research works, the original watermark image is not converted into ciphertext by any optical system and the entire watermark embedding & retrieval process is implemented fully digitally. However, the host image in these works is a digital hologram rather than a conventional photograph. The digital watermarking algorithms are designed according to the unique image characteristics of hologram images. These works are closely related to the topic of optical watermarking and are included in this paper as well.

In the acquisition, transmission and storage process, digital holograms can be damaged, corrupted or even maliciously tampered. The data integrity of digital holograms has to be guaranteed. Fragile watermark can be used to authenticate and check the hologram integrity since it is very sensitive to modifications in the host hologram.

In [64], one digital hologram is evenly divided into blocks and discrete cosine transform (DCT) is performed on each block. The binary watermark pattern is embedded by quantizing one DCT coefficient in each block and maintain all the other coefficients unchanged. The locations and quantized values of the DCT coefficients are determined by the watermark pattern pixel values. The experimental results indicate that a very small amount of bit errors in the hologram can be detected. To enhance the security, the DCT coefficients for watermark embedding are randomly selected by a pseudo random number (PRN) generator based on cellular automata [65], rather than deterministic locations [64].

In [66], the watermark is embedded in the Hadamard domain of hologram blocks instead of DCT domain and an improved quantization scheme is proposed. As an advantage, Hadamard transform has a lower computation complexity than Discrete Cosine Transform (DCT). In addition, the proposed watermarking scheme can support different watermarking capacities when the hologram resolution levels vary.

In addition to fragile hologram watermarking methods, some attempts have been made on watermarking a grayscale photograph image into a particular type of holograms, such as phase-only hologram and binary hologram. The problem of inconsistency in data representation format between gray scale image and phase-only image (or binary image) is addressed in these works with appropriate algorithms.

A phase only hologram (or kinoform) can be generated from a complex hologram with error diffusion method. In the error diffusion process (row by row scanning), certain bits in the phase value of each visited pixel [67] are replaced by the watermark data. The embedding bit locations for each hologram pixel can be further employed as an encryption key to enhance the watermarking security [68].

The embedding of a grayscale intensity image into a binary host hologram in a robust manner is investigated [69-71]. The grayscale image is encoded as binary bits by error diffusion based binarization [69], block truncation coding [70] or JPEG-BCH coding [71] and then embedded into a binary hologram by random scrambling. The preservation of host hologram quality after watermarking is one crucial issue in watermarking processing. MCA image inpainting algorithm can be employed to restore the host hologram for scrambling embedding watermarking [72].

In future works, the embedding algorithms for other types of ciphertexts (rather than holograms) and other types of host images (e.g. elemental image array (EIA) for integral imaging [73,74]) can be more extensively investigated, since they have some different characteristics compared with a hologram.

4. Conclusion

To the best of our knowledge, this is the first comprehensive summary of various optical architectures and related processing algorithms for optical image hiding or optical watermarking proposed in the past decade. The past research works are focused on two major aspects, various optical systems for image hiding and the methods for embedding optical system output into a host image. The previously proposed optical systems for image hiding include Double Random Phase Encoding (DRPE), off-axis holography system, phase shifting holography system, cascaded phase only mask architecture, Joint Transform Correlator (JTC), ghost imaging system and ptychography system. The features of each system are compared and discussed. Furthermore, different methods for embedding optical system output (ciphertext) into a host image are summarized, from simple weighted addition to sophisticated adaptive embedding. Some future research directions for optical image hiding are discussed in this paper.

3035-3046.